\documentclass{article}

\PassOptionsToPackage{numbers, compress}{natbib}

     \usepackage[preprint]{neurips_2021}

\usepackage[utf8]{inputenc} 
\usepackage[T1]{fontenc}    
\usepackage{hyperref}       
\usepackage{url}            
\usepackage{booktabs}       
\usepackage{amsfonts}       
\usepackage{nicefrac}       
\usepackage{microtype}      
\usepackage[ruled,vlined]{algorithm2e}
\usepackage{algorithmicx}
\usepackage{authblk}
\usepackage{graphicx}
\usepackage{subcaption}
\usepackage{mwe}
\setcitestyle{numbers}
\usepackage{enumitem}
\setenumerate[1]{label=(\arabic*)}
\usepackage{appendix}

\newcommand{\tss}{traditional summary statistics}

\title{Fair SA: Sensitivity Analysis for Fairness in Face Recognition}

\author{%
  Aparna R. Joshi  
  \quad
  Xavier Suau 
    \quad
  Nivedha Sivakumar 
    \quad
  Luca Zappella
  \quad
  Nicholas Apostoloff \\
  Apple \\
   \texttt{\{aparna\_joshi, xsuaucuadros, nivedha\_s, lzappella, napostoloff\}@apple.com} \\
}

\begin{document}

\maketitle

\begin{abstract}

As the use of deep learning in high impact domains becomes ubiquitous, it is increasingly important to assess the resilience of models. One such high impact domain is that of face recognition, with real world applications involving images affected by various degradations, such as motion blur or high exposure. Moreover, images captured across different attributes, such as gender and race, can also challenge the robustness of a face recognition algorithm. While traditional summary statistics suggest that the aggregate performance of face recognition models has continued to improve, these metrics do not directly measure the robustness or fairness of the models. Visual Psychophysics Sensitivity Analysis (VPSA) \cite{richardwebster2018visual} provides a way to pinpoint the individual causes of failure by way of introducing incremental perturbations in the data. However, perturbations may affect subgroups differently. In this paper, we propose a new fairness evaluation based on robustness in the form of a generic framework that extends VPSA. With this framework, we can analyze the ability of a model to perform fairly for different subgroups of a population affected by perturbations, and pinpoint the exact failure modes for a subgroup by measuring targeted robustness. 
With the increasing focus on the fairness of models, we use face recognition as an example application of our framework and propose to compactly visualize the fairness analysis of a model via AUC matrices. We analyze the performance of common face recognition models 
and empirically show that certain subgroups are at a disadvantage when images are perturbed, thereby uncovering trends that were not visible using the model's performance on subgroups without perturbations.

\end{abstract}
\section{Introduction}

Face recognition has been routinely used in private and government organizations to automate complex decision making processes \cite{sixta2020fairface}. The adoption of this technology into criminal justice systems \cite{davies2018evaluation, valentino2020police}, consumer applications, such as smartphones, and healthcare, heightens the potential negative implications that unfair algorithms can have on society. As cases of racial, gender and other biases surface, the inequity in face recognition algorithms have prompted calls for stricter regulation \cite{buolamwini2018gender, angwin2016machine, raji2020saving}. This inequity motivates both, the development of algorithms that treat sub-populations equally, and the need for companies to continually assess their technology for bias in order to reduce consumer harm.


The development of face recognition models relies on large scale datasets, and progress is often measured using summary statistics, such as accuracy, true positive rate, etc over the entire dataset (for brevity, we will refer to these statistics as {\em traditional} summary statistics). However, \tss{} do not provide enough information to explain individual failure modes, and thus provide little insight into how to diagnose and address issues in the models \cite{richardwebster2018visual}. Moreover, face recognition models can have classification accuracies as high as 99\% on benchmark datasets, but these models are often susceptible to failing under image degradations \cite{tong2021facesec}, and the outcomes can be dissimilar for different sub-populations \cite{buolamwini2018gender}. 
Performance of face recognition systems can be impacted by changes in the image processing pipeline (white balancing, enhancement, etc.), by artifacts related to the camera (noise, distortion, etc.) and other degradation artifacts due to motion such as motion blur or lighting conditions like exposure. 
Thus, along with the traditional evaluation of face recognition models, studying these algorithms with input images subject to perturbations can provide useful insights in evaluating robustness to visual distortions that can occur in real-world environments.


\begin{figure*}[t]
        \centering
        \begin{subfigure}[b]{0.32\textwidth}
            \centering
            \includegraphics[width=\textwidth,height=1.5in]{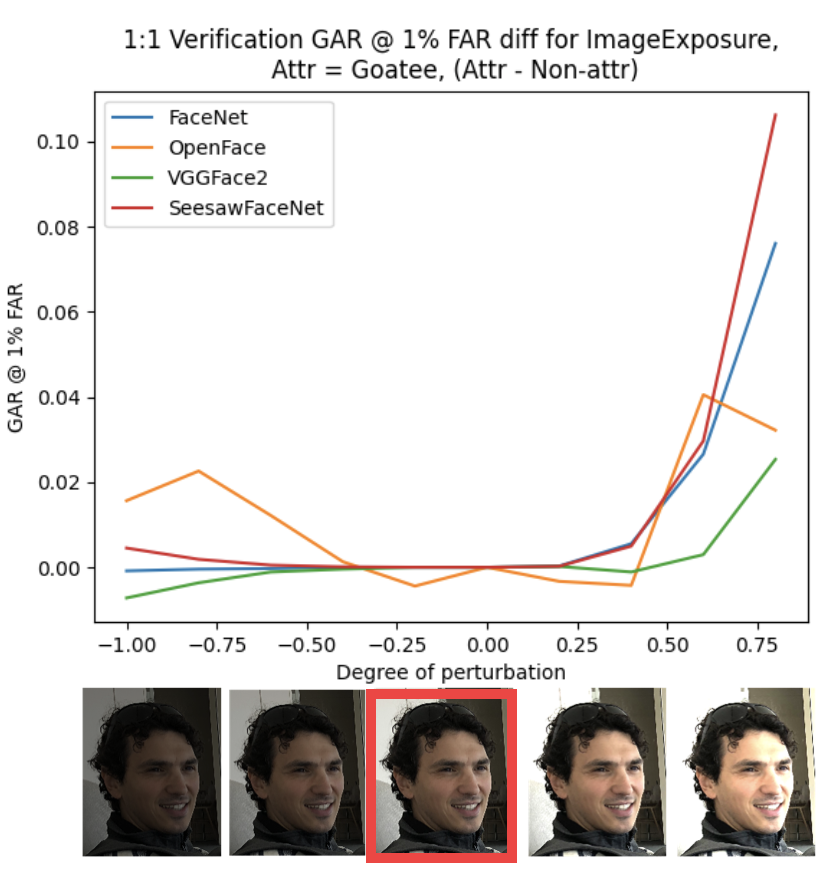}
            \caption{Protected: Goatee}%
            \label{fig:vd_a}
        \end{subfigure}
        \begin{subfigure}[b]{0.32\textwidth}  
            \centering 
            \includegraphics[width=\textwidth,height=1.5in]{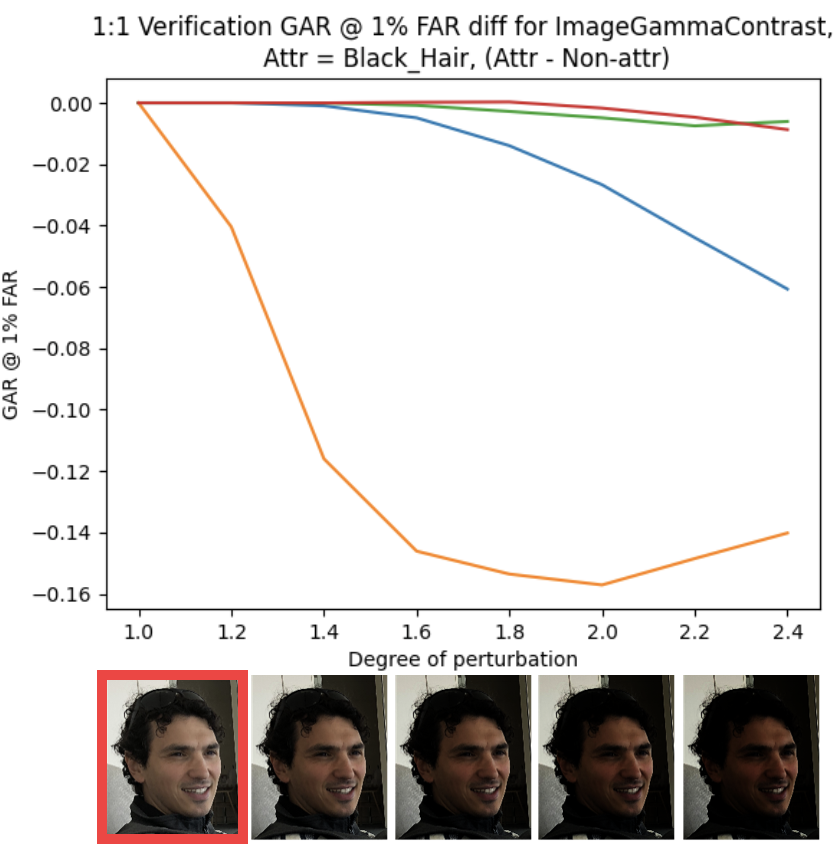}
            \caption{Protected: Black hair}%
            \label{fig:vd_b}
        \end{subfigure}
        \begin{subfigure}[b]{0.32\textwidth}  
            \centering 
            \includegraphics[width=\textwidth,height=1.5in]{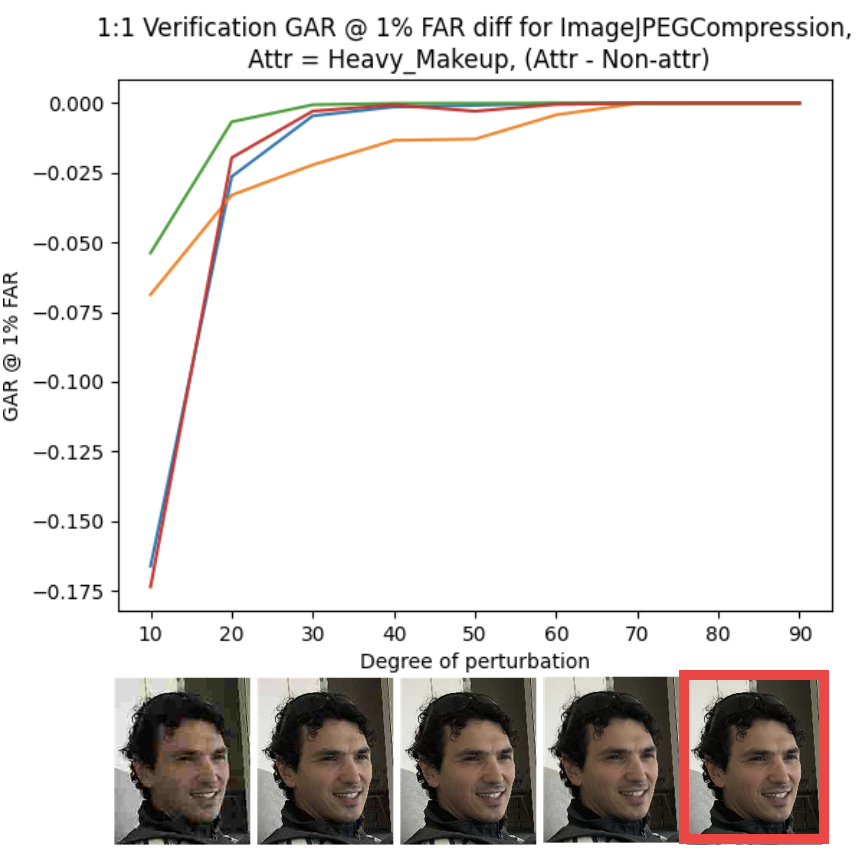}
            \caption{Protected: Heavy Makeup}%
            \label{fig:vd_c}
        \end{subfigure}
        \caption{Example Fair SA curves for face verification. In each plot, the original unperturbed image is highlighted. Models have a positive bias with increasing perturbation in Fig. \ref{fig:vd_a} favoring the \textit{goatee} subgroup under high exposure. Similarly, models disfavor \textit{black hair}  (Fig. \ref{fig:vd_b}) and \textit{heavy makeup}  (Fig. \ref{fig:vd_c}) subgroups (with negative bias values) under gamma contrast and JPEG compression respectively.}
        \label{fig:ver_diff}
    \end{figure*}

A number of works have shown divergent error rates of face recognition models across demographics \cite{buolamwini2018gender}.
This discrepancy in performance across sub-populations elevates the need to assess face recognition models for fairness. Many metrics have been proposed to assess the fairness of an algorithm \cite{du2020fairness}. One line of work studies \textit{individual fairness}, which relies on the notion that similar inputs should yield similar predictions \cite{dwork2012fairness}. On the other hand, \textit{group fairness} focuses on grouping examples based on a protected attribute, and measures whether the protected and the unprotected groups are treated similarly. Proposals for group fairness include metrics such as demographic parity, which requires the predictions of the model to be independent of the protected attribute. However, evaluation based on conventional fairness metrics does not account for robustness to image degradations, which frequently appear in real-world face recognition systems.

Fields studying vision have employed a technique called visual psychophysics to evaluate algorithms. Psychophysics is a set of experimental procedures from psychology and neuroscience that involve examination of relationships between perturbed test images and the responses they elicit in models \cite{webster2018psyphy}. 
The methodology described in \cite{richardwebster2018visual} (VPSA) derives ideas from visual psychophysics to measure the robustness of a face recognition model to incremental perturbations. Model performance is evaluated at various perturbation levels, providing explainability for failure cases beyond summary statistics. However, VPSA does not measure the robustness of the model with respect to fairness, since image degradations can affect different demographics differently.  For instance, blurred images may cause face recognition models to favor certain subgroups (e.g. young) over others (e.g. old). Such a discrepancy can be observed when face recognition is performed on images captured by an out-of-focus lens, or when there is relative motion between the object and the camera. 
In this paper, we make the following contributions:
\begin{enumerate}[leftmargin=*]
    \item  We propose Fair Sensitivity Analysis (Fair SA) (Fig. \ref{fig:ver_diff}) as a new way of measuring group fairness through robustness. Fair SA is an evaluation framework that allows machine learning practitioners to jointly evaluate the vulnerability of models to (i) image perturbations, and (ii) variations in demographics. 
    This approach allows us to analyze how a model treats different subgroups under perturbations of varying levels. 
    \item 
    The analysis of multiple models, across many attributes and perturbations generates a large amount of data. We propose the use of area under the curve (AUC) matrices in order to summarize the performance  of each model. AUC matrices provide a at-a-glance overview and can be efficiently analyzed row-, column-, and matrix-wise to extract explicit information about the bias induced by individual perturbations and the impact on individual subgroups.
    \item  We survey the performance of common face recognition models from the literature, such as VGGFace2 \cite{cao2018vggface2}, OpenFace \cite{amos2016openface}, FaceNet, \cite{schroff2015facenet} and SeesawFaceNet \cite{zhang2019seesawfacenets} on the widely used public dataset CelebA \cite{liu2018large} to demonstrate the efficacy of Fair SA. We evaluate the robustness of these algorithms to protected subgroups under varying levels of 9 types of perturbations on the tasks of verification and self-matching. Using this analysis, we uncover consistent biases against certain subgroups under specific data perturbations (e.g. the effect of \textit{exposure} on the \textit{pale skin} subgroup).
\end{enumerate}
\section{Related Work}

\textbf{Perturbed Inputs and Psychophysics ---} Several studies have explored simulating real world conditions that impact the performance of face recognition algorithms.  These simulations are often incremental perturbations that are applied to transform images. Karahan et al. \cite{karahan2016image} present an evaluation to assess the impact of incrementally perturbing face images on deep CNN face recognition models and found that while Gaussian blur, noise, and occlusion cause a significant decrease in performance, models were more robust to color distortions and changes in color balance. While these findings were for the closed set identification task, Grm et al. \cite{grm2018strengths} propose a similar evaluation framework for face verification, and conclude that noise, blur, and brightness greatly impact the performance of models, whereas the effects of compression and contrast are limited. 
As in VPSA, perturbations constitute an important component in our methodology.

Psychophysics studies the relationships between controlled stimuli and resulting model responses. 
Several fields in vision have looked at psychophysics as an alternate way of evaluating algorithms \cite{leibo2018psychlab}. Psyphy \cite{webster2018psyphy} is a comprehensive evaluation framework for object recognition models based on psychophysics and item-response theory. VPSA \cite{richardwebster2018visual} builds on Psyphy to support a similar item-response analysis for face recognition. VPSA comprises of a set of procedures developed to make face recognition algorithms more explainable. Fair SA, in turn, enhances VPSA to analyze the implications of variations in responses evoked for different data subgroups. These subgroups can be determined based on the sensitive attributes thereby yielding a fairness perspective to algorithmic robustness evaluation.

\textbf{Robustness ---} 
Robustness is particularly important for face recognition due to the inevitable influence of camera and real world conditions on the inputs. 
Studies have focused on evaluating models for robustness against digital attacks, such as variations in pose, illumination \cite{little2005methodology}, contrast and blur \cite{karahan2016image}, adversarial attacks \cite{yang2020delving}, and against physical attacks \cite{tong2021facesec}. Robustness against bias is a relatively recent area of research that has garnered much interest due to its widespread impact in the community. 
The idea of \textit{robustness bias}, which indicates that certain subgroups may be less robust to adversarial attacks, has been discussed in \cite{nanda2021fairness}. However, \cite{nanda2021fairness} proposes measuring the robustness bias by estimating the distance of subgroup members to the (closest) decision boundary relative to other members using bound approximations, and does not use perturbations to measure robustness. On the contrary, our method offers information on whether a model favors (disfavors) a subgroup under a perturbation, and can offer insights into potential mitigation strategies (e.g. informed training data augmentations). Thus, even though this experimental study shares an underlying motivation with this work, it is qualitatively and quantitatively different from the psychophysics-inspired approach we describe. 

\textbf{Fairness ---} Demographic differences in model performance have been studied over the past decades. In 2002, the seminal Face Recognition Vendor Test (FVRT) \cite{phillips2003face} already showed results signaling higher performance for males compared to females, and for older persons compared to younger persons. 
Apart from FRVT, there are several smaller scale evaluations of bias in trained algorithms on public datasets \cite{srinivas2019face, cavazos2020accuracy}.  
More recently, Buolamwini and Gebru \cite{buolamwini2018gender} analyzed commercial face recognition software on per-class accuracy for gender and skin tone classes, and demonstrated poorer performance of three commercial systems for dark skinned females than for lighter skinned males.
Khiyari et al. \cite{Khiyari2016FaceVS} measure the verification performance of VGGFace \cite{parkhi2015deep} on one, two, and three-class demographic groups based on race, gender and age with the best performance for middle-age white males and worst performance for young black females. 
The work in \cite{vangara2019characterizing} provides insights into how genuine and imposter score distributions differ between African-American and Caucasian image cohorts. 
The 2020 ChaLearn ``Looking at People'' challenge \cite{sixta2020fairface} summarizes the top submissions for evaluating accuracy and bias on gender and skin tones on the task of face verification. However, these studies do not study how performance varies across demographics for degraded input data. 

\section{Method}

VPSA \cite{richardwebster2018visual} applies visual psychophysics to assess face recognition models by introducing incremental perturbations, such as blur and rotation to input images. An example of each of the perturbations we consider is shown in Appendix \ref{sec.data}. Each of these perturbations is varied in magnitude thus controlling the difficulty of recognizing faces. VPSA assesses model robustness by comparing a perturbed image to itself (whilst ensuring it is sufficiently distinct from other images). Note that VPSA assumes one image per identity. It generates item-response curves (IRC) to quantify robustness as match rates (ratio of number of correct self-matches to the total number of images) on the y-axis against perturbation levels on the x-axis. Instead of using summary statistics to gauge performance, VPSA leverages visual psychophysics to compute these IRCs that map stimulus responses to performance, thereby allowing one to pinpoint the exact condition or stimulus type and level at which the algorithm can no longer reliably perform the task \cite{webster2018psyphy}.

Although VPSA does an excellent job at evaluating model robustness with respect to external adversaries and incremental degradations, it does not assess if a model is also robustly fair. Incremental degradations may affect a specific subgroup much more than the others (e.g. \textit{exposure} may treat certain skin tones unfairly). Therefore, we propose to enhance VPSA by enabling measurement of the effect that incremental perturbations can have on different demographics by way of computing the bias between protected and unprotected subgroups indicating if the algorithm favors (positive value) or disfavors (negative value) a specific population. This analysis can uncover discrepancies in model performance across subgroups under simulations of real world conditions, and gives us a way to carefully analyze these dissimilar behaviors to correct algorithmic deficiencies.

\begin{algorithm}[t]
\SetAlgoLined
\textbf{Input: } $D$, input image data  \\
\textbf{Input: } $t$, similarity threshold \\
\textbf{Input: } $n$, number of stimulus levels \\
\textbf{Input: } $b_l$ and $b_u$, the lower and upper bounds of stimulus levels \\
\textbf{Input: } $a$ and $v$, protected attribute and value \\
\textbf{Input: } $L$, identity labels  \\
\textbf{Input: } $M$, metadata (attribute annotations) \\
\textbf{Input: } $T$, perturbation type\\

    \begin{algorithmic}[1]
        \State $\Delta \leftarrow n $ stimulus levels from $b_l$ to $b_u$
        \State $ k \leftarrow \bigcup\limits_{\delta \in \Delta} {\phi (D, t, \delta, a, v, L, M, T)} $
    \end{algorithmic}

 \textbf{Output:} $k$, the Fair SA curve

 \caption{$C(D, t, n , b_l, b_u, a, v, L, M, T)$: Fair SA curve generation function for perturbation type $T$}
 \label{alg:alg1}
\end{algorithm}


\begin{algorithm}[t]
\SetAlgoLined
\textbf{Input: } $D$, input image data \\
\textbf{Input: } $t$, similarity threshold \\
\textbf{Input: } $\delta$, stimulus level \\
\textbf{Input: } $a$ and $v$, protected attribute and value \\
\textbf{Input: } $L$, identity labels \\
\textbf{Input: } $M$, metadata (attribute annotations) \\
\textbf{Input: } $T$, perturbation type\\

    \begin{algorithmic}[1]
        \State $ D' \leftarrow i \in D : T(i, \delta) $
        \State $pred \leftarrow  \gamma (D', D, t)$
        \State $\alpha \leftarrow \beta(a, v, D, M, L, pred)$
    \end{algorithmic}

 \textbf{Output:} $\{\delta, \alpha\}$, an $x$,$y$ coordinate pair (stimulus level, bias)

 \caption{$\phi (D, t, \delta, a, v, L, M, T)$: Fair SA point generation function for stimuli $T(i, \delta)$}
 \label{alg:alg2}
\end{algorithm}

 VPSA begins by pruning the identities that lead to false matches or false non-matches for a face recognition algorithm. By retaining only the identities that match well to themselves and poorly to others, the effect of applied perturbations on the remaining input images is disentangled from model failures at no perturbation. On the other hand, Fair SA treats this initial pruning step as optional due to the possibility that the pruning may result in the removal of more identities belonging to some subgroups than others, which can adversely affect our fairness assessment. However, for self-matching, experiments with the original VPSA pruning step have also been performed and presented in Appendix \ref{sec.self_matching}. In our experiments, we observe similar trends for VPSA with and without the pruning step. For Fair SA with pruning, we largely observe the same signedness albeit different magnitudes for the attributes as without pruning.


\begin{algorithm}[t]
\SetAlgoLined
\textbf{Input: } $f$, a function to extract model responses (representations from selected layers) from a face recognition model\\
\textbf{Input: } $D_p$, probe (perturbed) data \\
\textbf{Input: } $D_g$, gallery (original) data \\
\textbf{Input: } $t$, similarity threshold \\

     \begin{algorithmic}[1]
    \State $ R' \leftarrow i \in D_p : f(i) $   
          \State $ R \leftarrow i \in D_g : f(i) $ 
          \State $ S \leftarrow r' \in R', r \in R : dist(r, r') $
          \end{algorithmic}
    \uIf{TASK is face verification}{
          $ pred \leftarrow S $
    }
    \uElseIf{TASK is self-matching}{
     $M \leftarrow S \geq t$ \\
     $pred \leftarrow diag(M)$
    }
 \textbf{Output:} $pred$, the task-dependent predictions.
 
 \caption{$\gamma(D_p, D_g, t)$: function that generates task-dependent predictions for face recognition model $f$}
 \label{alg:alg3}
\end{algorithm}


\begin{algorithm}[t]
\SetAlgoLined
\textbf{Input: } $b$, function to compute bias w.r.t. protected attribute \\
\textbf{Input: } $a$ and $v$, protected attribute and value \\
\textbf{Input: } $D$, input image data \\
\textbf{Input: } $M$, metadata (attribute annotations) \\
\textbf{Input: } $L$, identity labels \\
\textbf{Input: } $pred$, set of task-dependent predictions for the data \\

    \begin{algorithmic}[1]
        \State $P_{a, v} \leftarrow i \in D$ s.t. $M_i [a] = v$
        \State $P_{a, \bar{v}} \leftarrow D \setminus P_{a, v}$
        
                \State $\alpha \leftarrow b(pred_i | i \in P_{a, v}, pred_i | i \in P_{a, \bar{v}}, L) $  \Comment{Note: for face verification we have $pred_{i, j} | i,j \in P_a$}
    \end{algorithmic}

 \textbf{Output:} $\alpha$, the bias.
 
 \caption{$\beta(a, v, D, M, L, pred)$: Bias generation function for protected attribute $a$ and value $v$}
 \label{alg:alg4}
\end{algorithm}

Let $D$ be original input data, $a$ and $v$ be the protected attributes and values (e.g. $a$ can be the attribute \textit{gender}, and $v$ can be the value \textit{female}). Let $P_{a, v} \in D$ be the protected subgroup of the population (as defined by $a$ and $v$), and $P_{a, \bar{v}} = D \setminus P_{a, v}$ be the unprotected subgroup.
The Fair SA algorithm starts with function $C$ (Alg. \ref{alg:alg1}) that generates the final Fair SA curves in the form of item-response values. $C$ is called once for each perturbation type $T$ and it repeatedly calls the function $\phi$ (Alg. \ref{alg:alg2}) for stimulus levels ranging from $b_l$ to $b_u$ for each input perturbation type. Function $\phi$ generates a point on the Fair SA curve, where the point represents the bias $\alpha$ of the face recognition algorithm $f$ on protected subgroup $P_{a, v}$. A bias measure can be task-dependent e.g. for face verification, we use Genuine Acceptance Rate (GAR) at 1\% False Acceptance Rate (FAR) \cite{singh2019recognizing}, and for self-matching, we use statistical imparity.

The perturbation function $T$ in Alg. \ref{alg:alg2} produces perturbed images given the original images and stimulus level $\delta$. In this paper, we refer to the perturbed images as the probes ($D_p=D'$), and the original images as the gallery ($D_g=D$). Similar to VPSA, the $\gamma$ function (Alg. \ref{alg:alg3}) is a wrapper to a given face recognition model $f$ that generates task-dependent model predictions. $\gamma$ computes cosine similarities between model responses (of a layer) for the probe set $D_p$, and the gallery set $D_g$.  For verification, we use the subgroup similarities and ground truth subgroup identities to compute Genuine Acceptance Rate (GAR) at 1\% False Acceptance Rate (FAR).
For self-matching, the predictions are the similarities of perturbed images with corresponding original images (e.g. diagonal of the thresholded similarity matrix). $\beta$ (Alg. \ref{alg:alg4}) then computes the statistical imparity, the difference between the probability that a random sample drawn from $P_{a, v}$ is a match with its perturbed version and the probability that a random sample from the complement $P_{a, \bar{v}}$ is a match.

\textbf{AUC Matrices.} Drawing conclusions about multiple models across different perturbations using curves generated by Alg. \ref{alg:alg1} can be challenging. Therefore, we propose the use of AUC matrices for each model to visualize Fair SA (or VPSA) results. Furthermore, we use matrix norms to understand the average model fairness, and row-wise or column-wise analysis to extract further information about fairness with respect to certain perturbations or subgroups. For Fair SA (Fig. \ref{fig:veri_celeba}), an element in the AUC matrix depicts (for a model), the signed AUC of the Fair SA curves\footnote{Similarly, for VPSA (Fig. \ref{fig:vanilla}), an element in the AUC matrix indicates the AUC of the IRC generated by VPSA for a particular perturbation.}. We propose to use the L1 norm to marginalize the AUC matrix, row, column or matrix-wise.

\section{Experiments}

We perform experiments on two tasks: face verification \cite{singh2019recognizing} (comparing a perturbed image with original images to verify whether a perturbed-original image pair belongs to the same identity), and self-matching (matching a perturbed image to its original image). We use the following pre-trained models without additional fine-tuning: VGGFace2 \cite{cao2018vggface2}, FaceNet \cite{schroff2015facenet}, OpenFace, \cite{amos2016openface} and SeesawFaceNet \cite{zhang2019seesawfacenets}. For all the models, we obtain responses from one of the final feature layers. 

\textbf{Data and perturbations.} We investigate model performances on the following 9 perturbations: \textit{Gaussian blur}, \textit{gamma contrast}, \textit{rotation}, \textit{speckle noise}, \textit{exposure}, \textit{saturation}, \textit{motion blur}, \textit{JPEG compression}, and \textit{vignette} (Examples in Appendix \ref{sec.data}). We evaluate our models on the CelebA test set which consists of $\sim$20k samples. Each sample has an identity label, and binary annotations for 40 attributes e.g. \textit{male}, \textit{wavy hair}, etc. 

\begin{figure}[t]
\centering
\includegraphics[width=0.8\textwidth, height=1.2in]{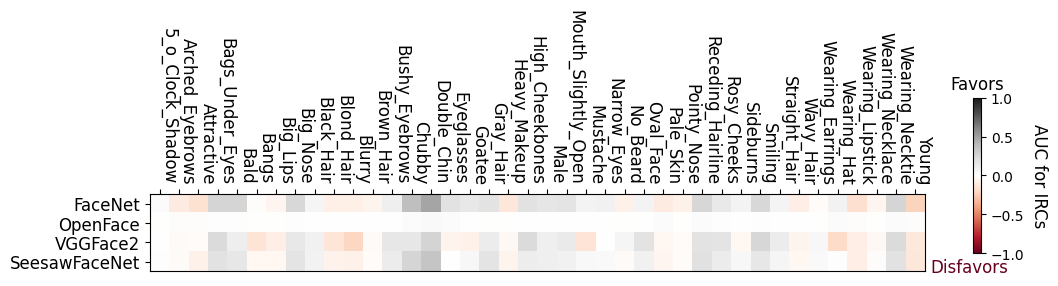}
\caption{Depiction of GAR difference at 1\% FAR between protected and unprotected subgroups at no perturbation. }%
\label{fig:fsa_verification_no_perturbation}
\end{figure}
    
\begin{figure}[t]
\centering
\includegraphics[width=0.9\textwidth, height=2.5in]{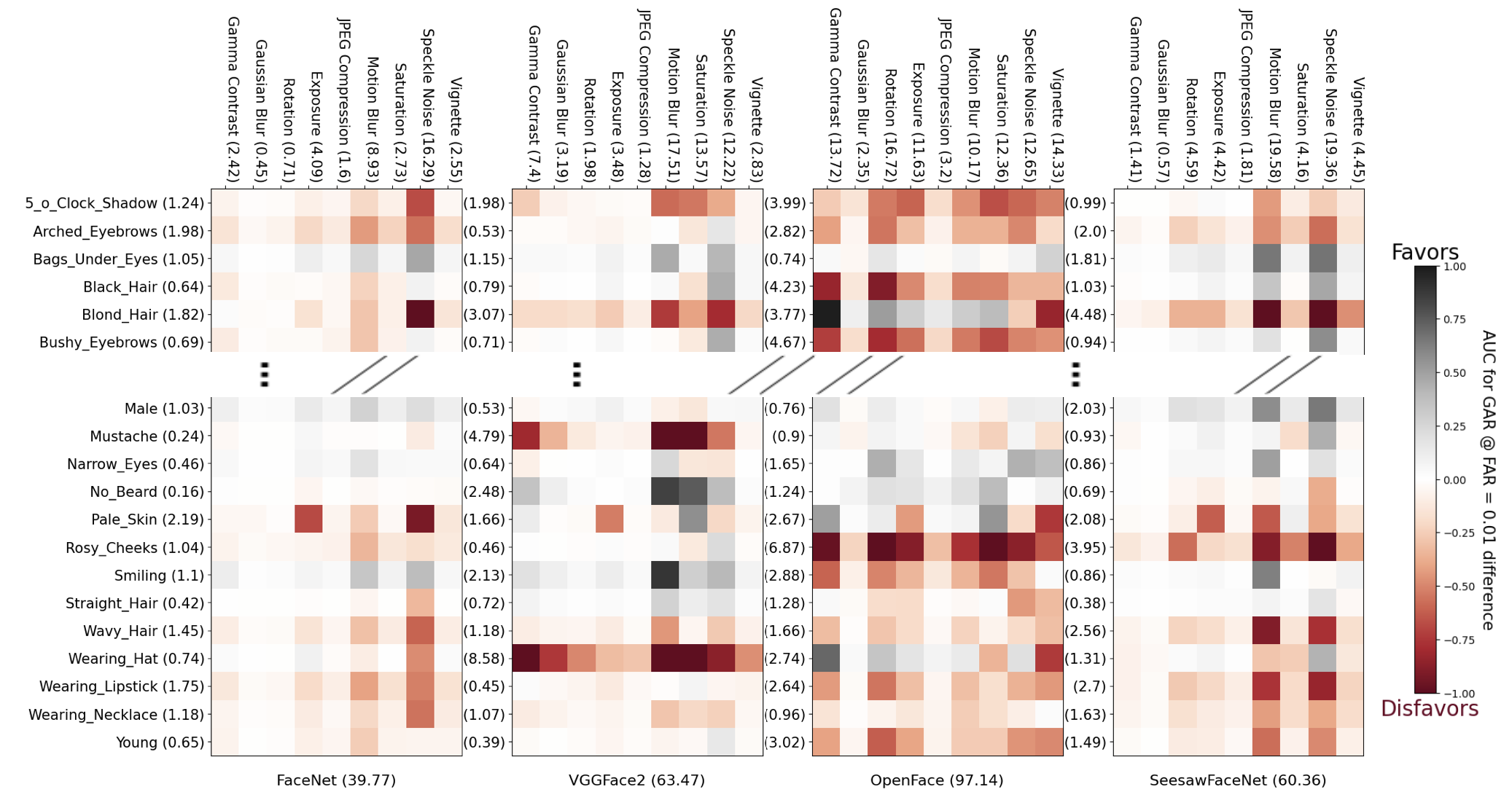}
\caption{Depiction of AUC for GAR difference at 1\% FAR for selected subgroups. Note that the numbers in attribute labels denote the attribute norms. 
Matrix L1 norms show that, on average, FaceNet is the most fair (lowest norm), and OpenFace is the least fair model. }%
\label{fig:veri_celeba}
\end{figure}

\textbf{Verification. } This task involves comparing the model responses for every perturbed image with every original image of all identities. A pair is a match if both images belong to the same identity. The overall performances of the models are evaluated in terms of Genuine Acceptance Rate (GAR) at False Acceptance Rate (FAR) \cite{singh2019recognizing} (Appendix \ref{sec.verification}, Fig. \ref{fig:verification} shows that VGGFace2 performs the best and OpenFace the worst at no perturbation). We consider the GAR at 1\% FAR metric which is commonly used for verification \cite{singh2019recognizing}, and plot Fair SA curves representing the differences between GAR at 1\% FAR across protected and unprotected subgroups at varying levels of perturbation (Fig. \ref{fig:ver_diff}). For Fair SA verification experiments, we implement a pruning procedure that eliminates the pairs of images that are false matches at no perturbation, so that the GAR at 1 \% FAR metric aligns for the protected and unprotected subgroups and we can use AUC to compare models. The smaller the absolute value of AUC for the Fair SA curves, the more fair is the algorithm. Note that we allow the AUC in Fair SA curves to be negative. In such cases, the unprotected subgroup is favored over the protected one. Fig. \ref{fig:fsa_verification_no_perturbation} indicates that OpenFace is the fairest model at no perturbation, with the lowest GAR difference at 1\% FAR. 
However, matrix L1 norms on the Fair SA AUC model matrices\footnote{see complete verification results and L1 norms in Appendix \ref{sec.verification}, Fig. \ref{fig:fsa_ver_all}} (Fig. \ref{fig:veri_celeba}) indicate that OpenFace is the least robustly fair model (based on the Fair SA score). Interestingly, while we see that at no perturbation, models (e.g. VGGFace2 and SeesawFaceNet) are fair with respect to different subgroups (e.g. \textit{pale skin}) (Fig. \ref{fig:fsa_verification_no_perturbation}), we see in Fig. \ref{fig:veri_celeba} that the models can disfavor those subgroups significantly in the presence of certain perturbations (e.g. \textit{exposure}). 

\begin{figure*}[t]
        \centering
        \begin{subfigure}[b]{0.32\textwidth}
            \centering
            \includegraphics[width=\textwidth,height=1.5in]{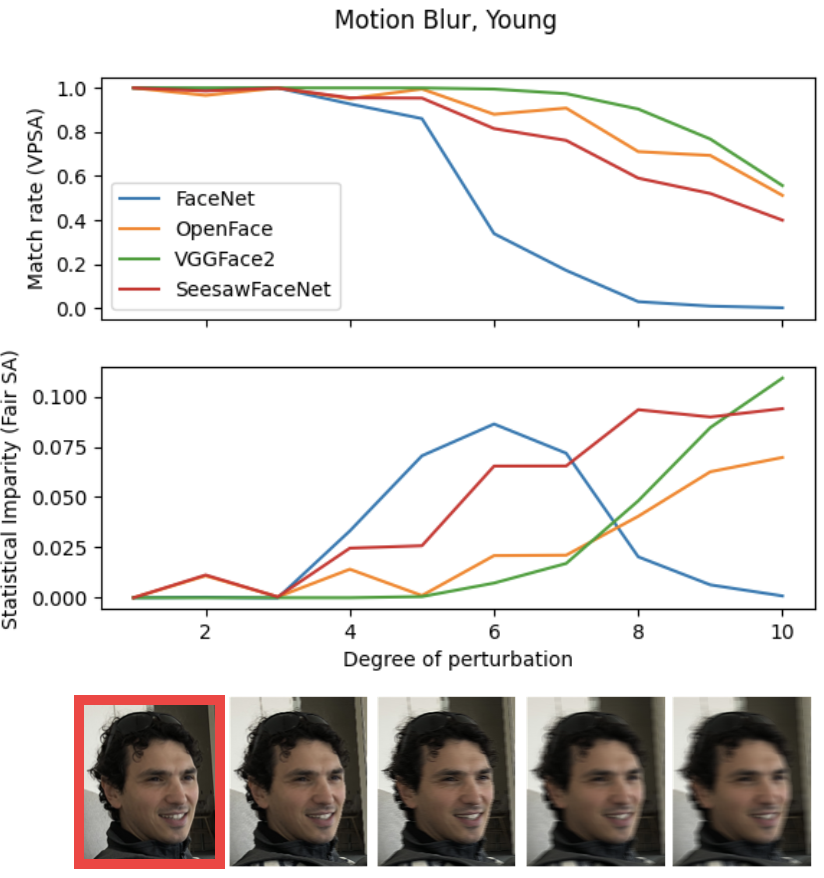}
            \caption{Protected: `Young'}%
            \label{fig:fsa_a}
        \end{subfigure}
        \begin{subfigure}[b]{0.32\textwidth}  
            \centering 
            \includegraphics[width=\textwidth,height=1.5in]{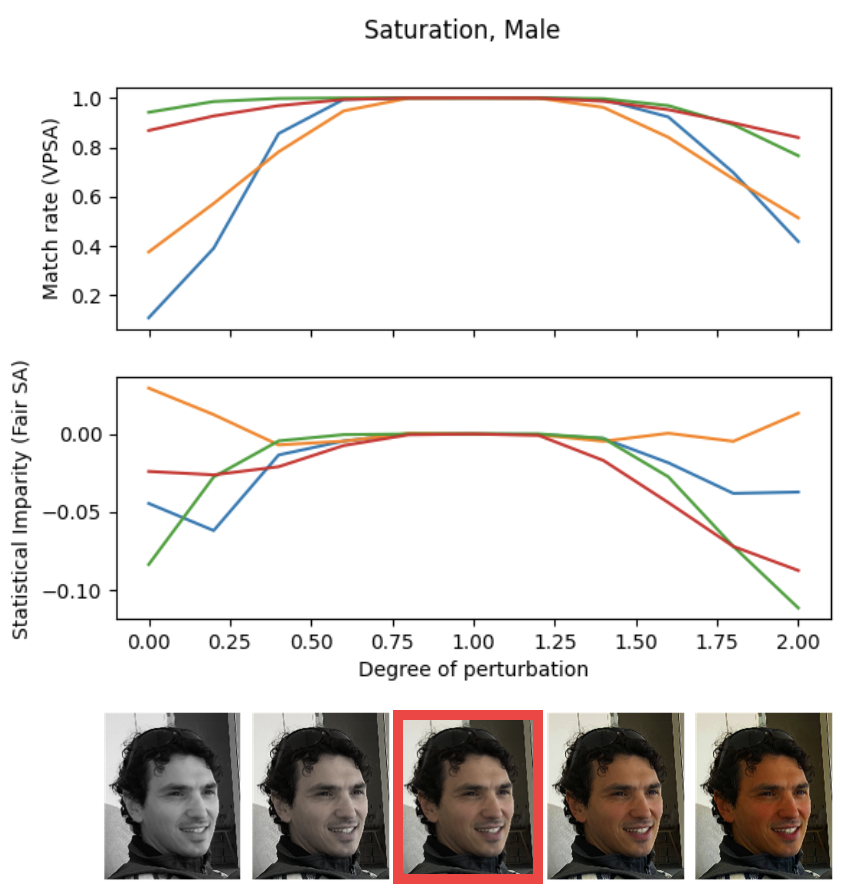}
            \caption{Protected: `Male'}%
            \label{fig:fsa_b}
        \end{subfigure}
        \begin{subfigure}[b]{0.32\textwidth}  
            \centering 
            \includegraphics[width=\textwidth,height=1.5in]{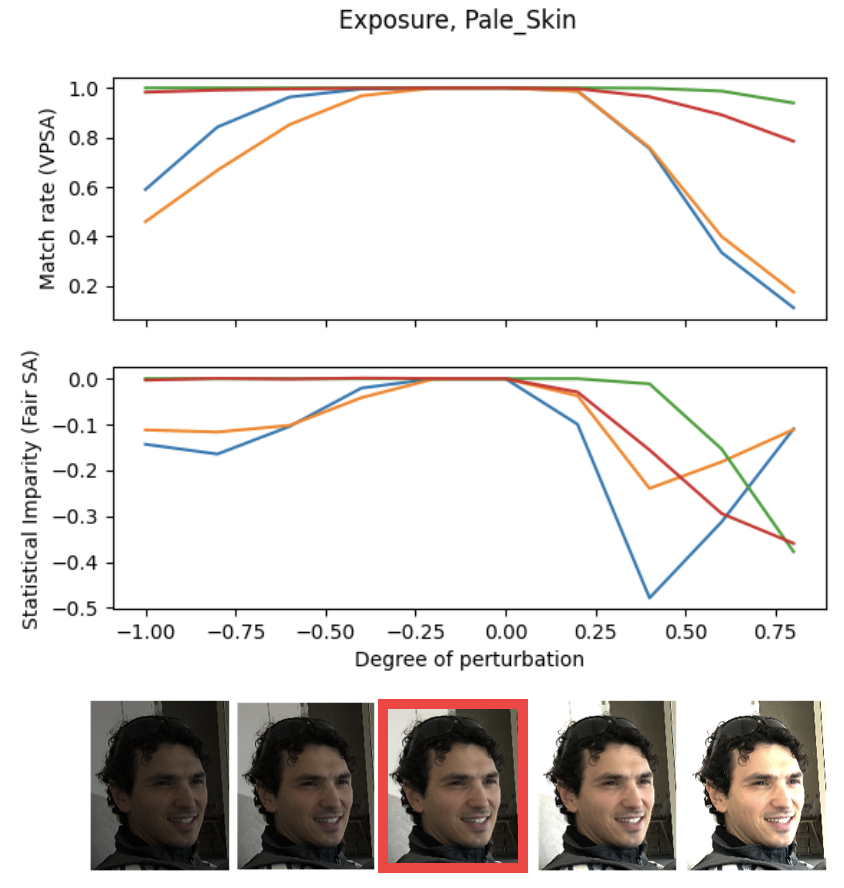}
            \caption{Protected: `Pale Skin'}%
            \label{fig:fsa_c}
        \end{subfigure}
        \caption{VPSA (top) and Fair SA (bottom) curves for self-matching. VPSA curves show that VGGFace2 is the most robust across \textit{motion blur}, \textit{saturation}, and \textit{exposure} perturbations. However, Fair SA curves show that the models are biased upon subjecting the data to perturbations e.g., all the models are above 0 in Fig. \ref{fig:fsa_a}, indicating that they favor the \textit{young} subgroup when subjected to \textit{motion blur}. Similarly, in the case of low and high \textit{saturation} (Fig. \ref{fig:fsa_b}) and \textit{exposure} (Fig. \ref{fig:fsa_c}), Fair SA indicates that models largely disfavor the \textit{male}, and \textit{pale skin} subgroups, respectively. Contrary to VPSA, Fair SA shows that at the highest level of perturbation, VGGFace2 is the least fair model.}
        \label{fig:fsa}
    \end{figure*}

\begin{figure}[t]
\centering
\includegraphics[width=0.9\textwidth, height=1.9in]{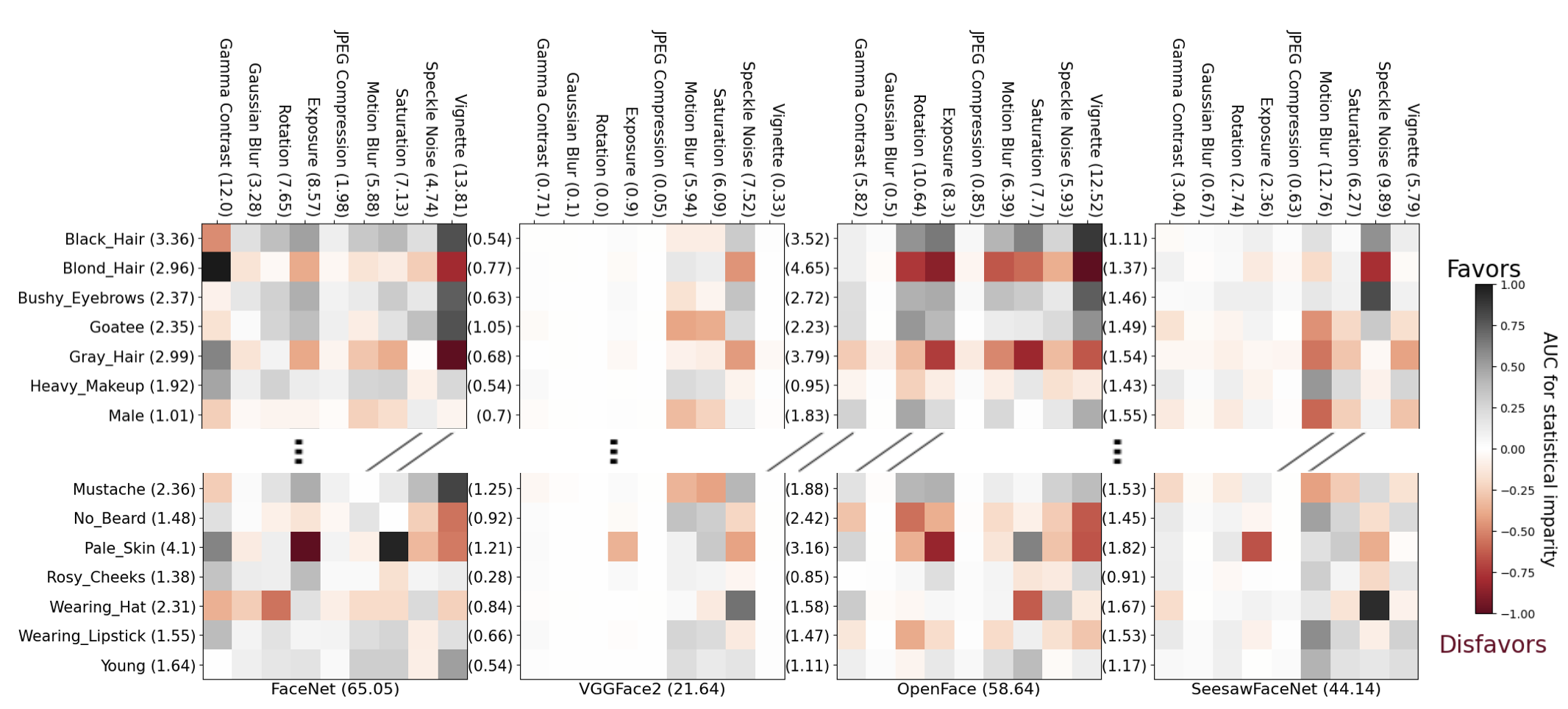}
\caption{Depiction of AUC for Fair SA curves (self-matching) for selected attributes. Note that the numbers in attribute labels denote the attribute L1 norms. 
Matrix L1 norms indicate that, on average, VGGFace2 is the most fair (with the lowest L1 norm), and FaceNet the least fair model. }%
\label{fig:sm_auc}
\end{figure}

\textbf{Self-matching.} This task is more generic in that it ignores the identities for the inputs, and can be used when a model is intended to be used for multiple tasks. It compares the model response for a perturbed image with that of the corresponding original image. VPSA shows that VGGFace2 largely outperforms the other models in robustness by withstanding perturbations to a greater degree (Fig. \ref{fig:fsa}). Matrix L1 norms on the Fair SA AUC model matrices\footnote{see complete self-matching results and L1 norms in Appendix \ref{sec.self_matching}, Fig. \ref{fig:auc_matrix_wopru_fsa}} (Fig. \ref{fig:sm_auc}) confirm that VGGFace2 is the most robustly fair model.  However, from a more detailed analysis, we can see that at the highest levels of perturbations like \textit{motion blur}, \textit{saturation}, and \textit{exposure}, VGGFace2 is the least fair with respect to certain subgroups (\textit{young}, \textit{male}, \textit{pale skin}). 
Note from Fig. \ref{fig:fsa_a} that Fair SA analysis on FaceNet indicates that with increasing \textit{motion blur}, first the bias increases and then decreases. However, the fairness of FaceNet at higher perturbation is likely due to the fact that the performance of FaceNet is drastically degraded for strong motion blur. This indicates that fairness should be analyzed in conjunction with robustness for a complete outlook on the models. 

\textbf{Row- and column-wise analysis.} Such an analysis can help with a broad and quick evaluation of the impact on subgroups and effect of perturbations. Across models for verification, the row-wise L1 norms indicate that models are most biased against \textit{wearing hat}, \textit{blond hair}, and \textit{rosy cheeks} (Fig. \ref{fig:veri_celeba}). Similarly, column-wise L1 norms across models indicate that the perturbations that most affect the model fairness are \textit{speckle noise}, \textit{motion blur}, and \textit{saturation}. Note that this analysis can also be performed per model to help understand the impact on subgroups and effect of perturbations for individual models.

\textbf{Effect of perturbations on salient features.} Unsurprisingly, perturbations that degrade salient details adversely affect the performance on those sub-groups that have more salient features than those that do not. In the case of self-matching, increasing \textit{Gaussian blur}, \textit{motion blur}, or \textit{JPEG compression} results in models favoring the \textit{young} attribute. We hypothesize that blurring images of faces of older people may result in loss of details (that are not as prominent in the faces of young people) that are important for face recognition. Some observations are common to both self-matching and verification tasks, such as, increasing image \textit{exposure} results in models favoring dark features, such as \textit{black hair}, \textit{bushy eyebrows}, and disfavoring features such as \textit{pale skin}.

\section{Conclusion}
We propose an evaluation framework called Fair SA to measure group fairness in the presence of data perturbations and demonstrate its use on two tasks: verification and self-matching.
An absolutely fair algorithm will have performance that is even on (all pairs of) protected and unprotected subgroups for incremental perturbations, and unfairness is any deviation from this fair algorithmic behavior. We provide experimental evidence that this unfairness can exist in some commonly used models trained on real world datasets, and we propose to visualize this unfairness via AUC matrices. Such matrices can be marginalized row and column wise to provide insights with respect to attributes or input perturbations. We show that even if models exhibit performance equity between protected and unprotected subgroups when there is no data degradation, in the presence of perturbations the models may start to favor (or disfavor) certain data subgroups. 
In future works, we will explore how Fair SA can be utilized to mitigate biases via targeted training data augmentation driven by the type and the magnitude of perturbations that cause biased behavior.


\clearpage

\clearpage

\bibliographystyle{unsrt}
\bibliography{main}
\clearpage
\appendix
\appendixpage

\section{Data and perturbations}\label{sec.data}
In this section, the perturbations applied on a sample image are presented. 

\begin{figure}[htbp]
\centering
\begin{tabular}{cccccc}
\subcaptionbox{Original\label{1}}{\includegraphics[width = 0.7in]{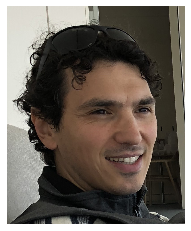}} &
\subcaptionbox{Gaussian blur\label{2}}{\includegraphics[width = 0.7in]{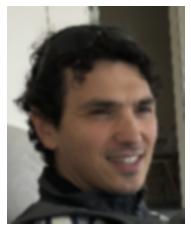}} &
\subcaptionbox{Gamma contrast\label{3}}{\includegraphics[width = 0.7in]{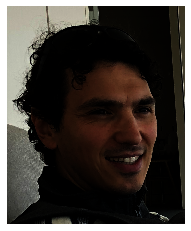}} &
\subcaptionbox{Rotation\label{4}}{\includegraphics[width = 0.7in]{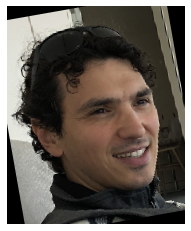}} &
\subcaptionbox{Speckle noise\label{5}}{\includegraphics[width = 0.7in]{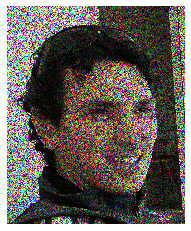}} &
\subcaptionbox{Motion blur\label{6}}{\includegraphics[width = 0.7in]{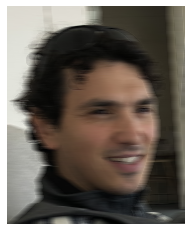}} \\
\subcaptionbox{JPEG compression\label{7}}{\includegraphics[width = 0.7in]{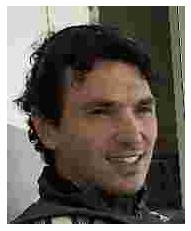}} &
\subcaptionbox{Vignette\label{8}}{\includegraphics[width = 0.7in]{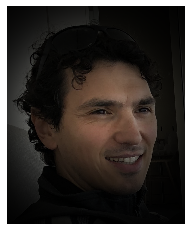}} &
\subcaptionbox{Low exposure\label{9}}{\includegraphics[width = 0.7in]{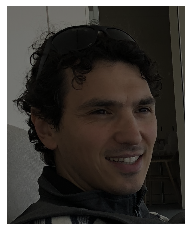}} &
\subcaptionbox{High exposure\label{10}}{\includegraphics[width = 0.7in]{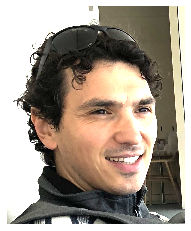}} &
\subcaptionbox{Low saturation\label{11}}{\includegraphics[width = 0.7in]{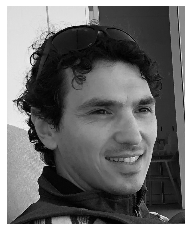}} &
\subcaptionbox{High saturation\label{12}}{\includegraphics[width = 0.7in]{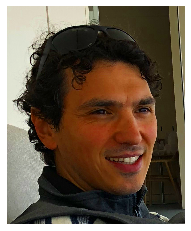}} \\
\end{tabular}
\caption{Perturbations applied to a sample image \ref{1}. For exposure and saturation, the range is varied from low to high.}
\label{perts}
\end{figure}

\section{Verification}\label{sec.verification}
\begin{figure}[htbp]
\centering
\includegraphics[width=0.65\textwidth, height=2.5in]{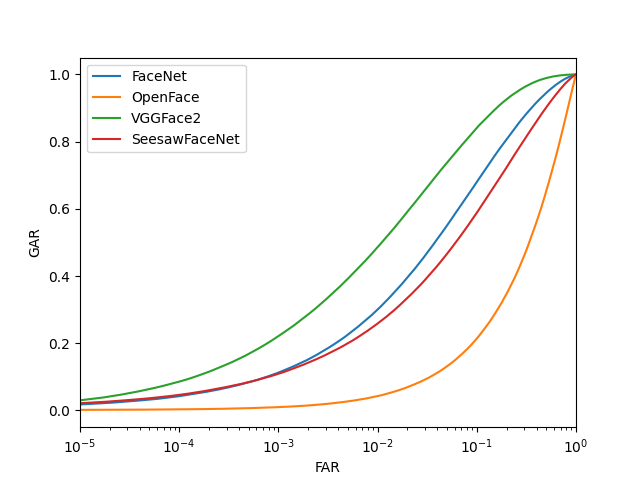}
\caption{Genuine Acceptance Rate (GAR) at False Acceptance Rate (FAR) for verification on CelebA at no perturbation. Higher the AUC, the better the performance of the model. Thus, VGGFace2 performs the best, and OpenFace the worst amongst the models compared.}%
\label{fig:verification}
\centering
\end{figure}

\begin{figure}[htbp]
\centering
\includegraphics[width=\textwidth, height=4in]{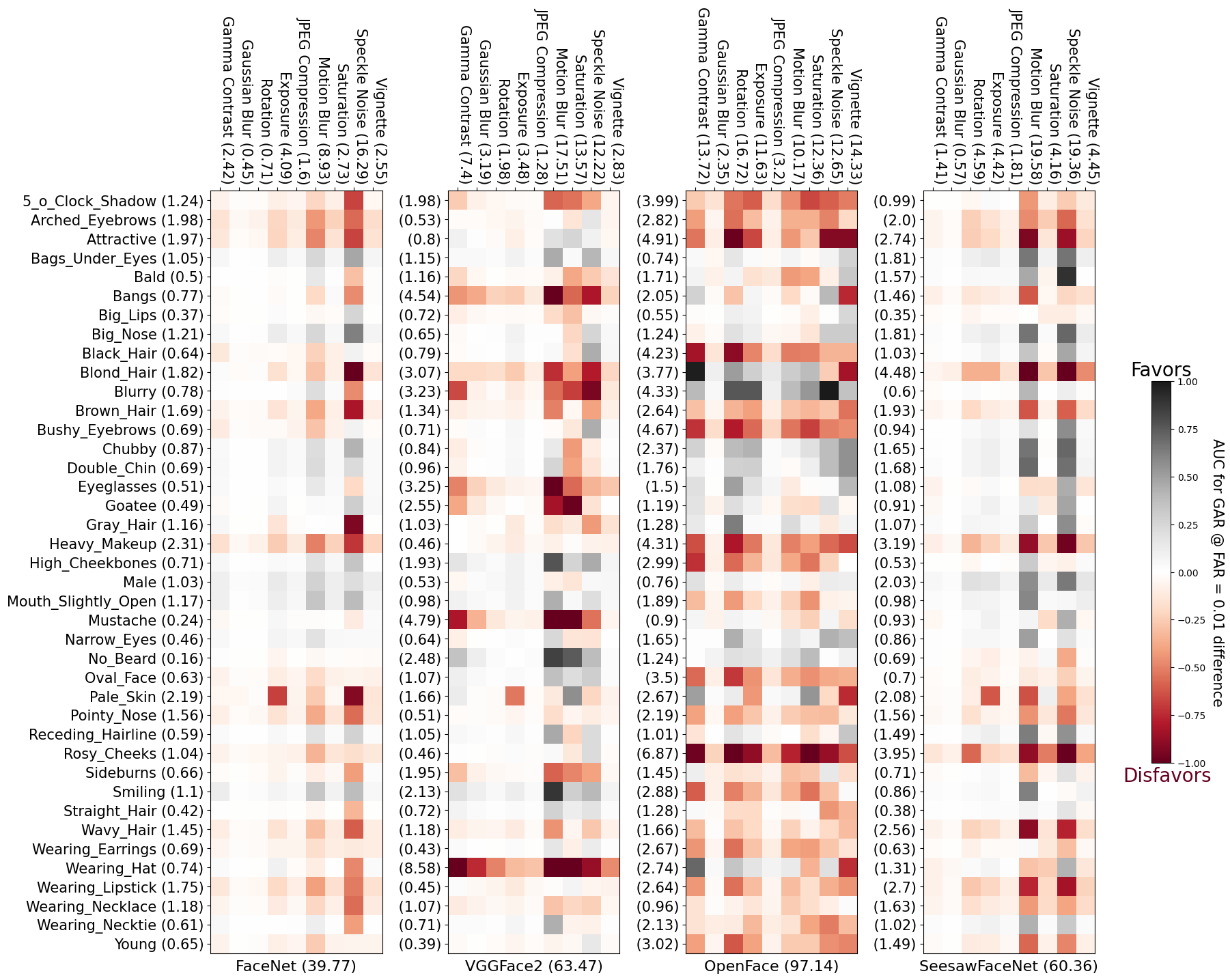}
\caption{Depiction of AUC for difference between GAR @ 1\% FAR values for all protected and unprotected subgroups. }%
\label{fig:fsa_ver_all}
\centering
\end{figure}

In this section, additional results for verification are presented. Fig. \ref{fig:verification} shows the GAR at FAR curves for different models at no perturbation. Fig. \ref{fig:fsa_ver_all} depicts the AUC matrices for all attributes.

\section{Self-matching}\label{sec.self_matching}
\begin{figure*}[htbp]
        \centering
        \begin{subfigure}[b]{0.45\textwidth}
            \centering
            \includegraphics[width=\textwidth,height=1.6in]{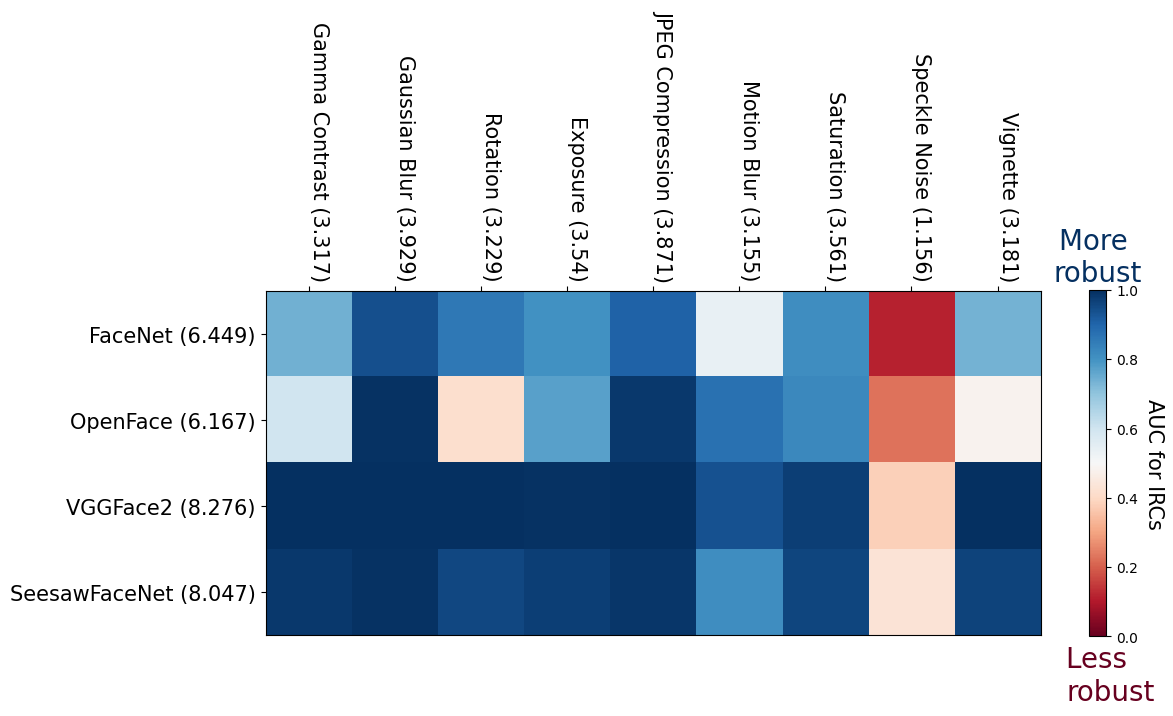}
            \caption{AUC matrix for VPSA without pruning}%
            \label{fig:vanilla_a}
        \end{subfigure}
        \begin{subfigure}[b]{0.45\textwidth}  
            \centering 
            \includegraphics[width=\textwidth,height=1.6in]{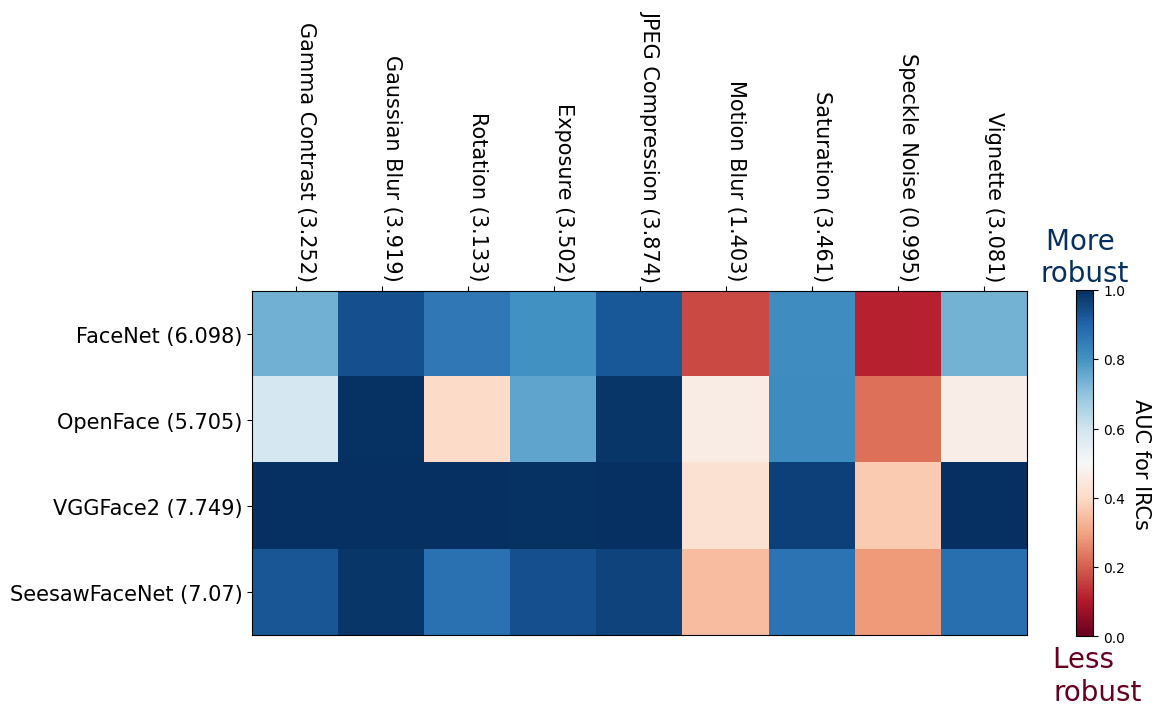}
            \caption{AUC matrix for VPSA with pruning}%
            \label{fig:vanilla_b}
        \end{subfigure}
        \caption{Fig. \ref{fig:vanilla_a} Depiction of AUCs for IRCs as computed by VPSA without pruning. Note that the numbers in the labels denote the L1 norms for the models and perturbations. Computing L1 norms for the rows indicates that, on average, VGGFace2 is the most robust (with the highest L1 norm), and OpenFace the least robust amongst the models compared.  Fig. \ref{fig:vanilla_b} Depiction of AUCs for IRCs as computed by VPSA with pruning. Trends observed are largely similar to those obtained without pruning. Computing L1 norms for the rows also indicates the same order of robustness of the models i.e., on average, VGGFace2 is the most robust (with the highest L1 norm), and OpenFace the least robust amongst the models compared, on CelebA.}
        \label{fig:vanilla}
    \end{figure*}

\begin{figure}[htbp]
\centering
\includegraphics[width=\textwidth, height=2.7in]{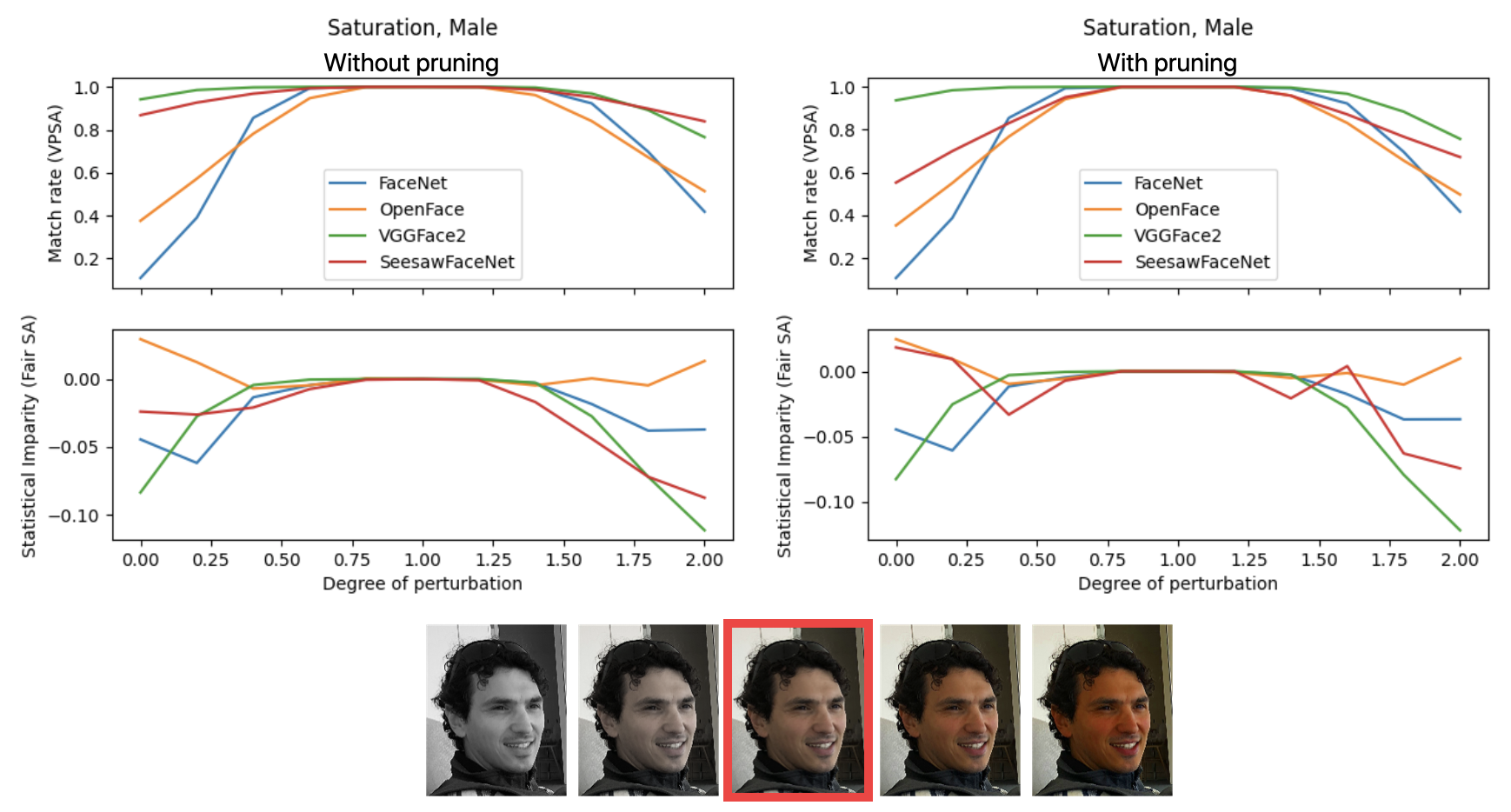}
\caption{Example VPSA and Fair SA curves without and with pruning for saturation with \textit{male} as the protected subgroup. We observe similar trends for individual models for without and with pruning.}%
\label{fig:compare}
\centering
\end{figure}

\begin{figure}[htbp]
\centering
\includegraphics[width=\textwidth, height=4in]{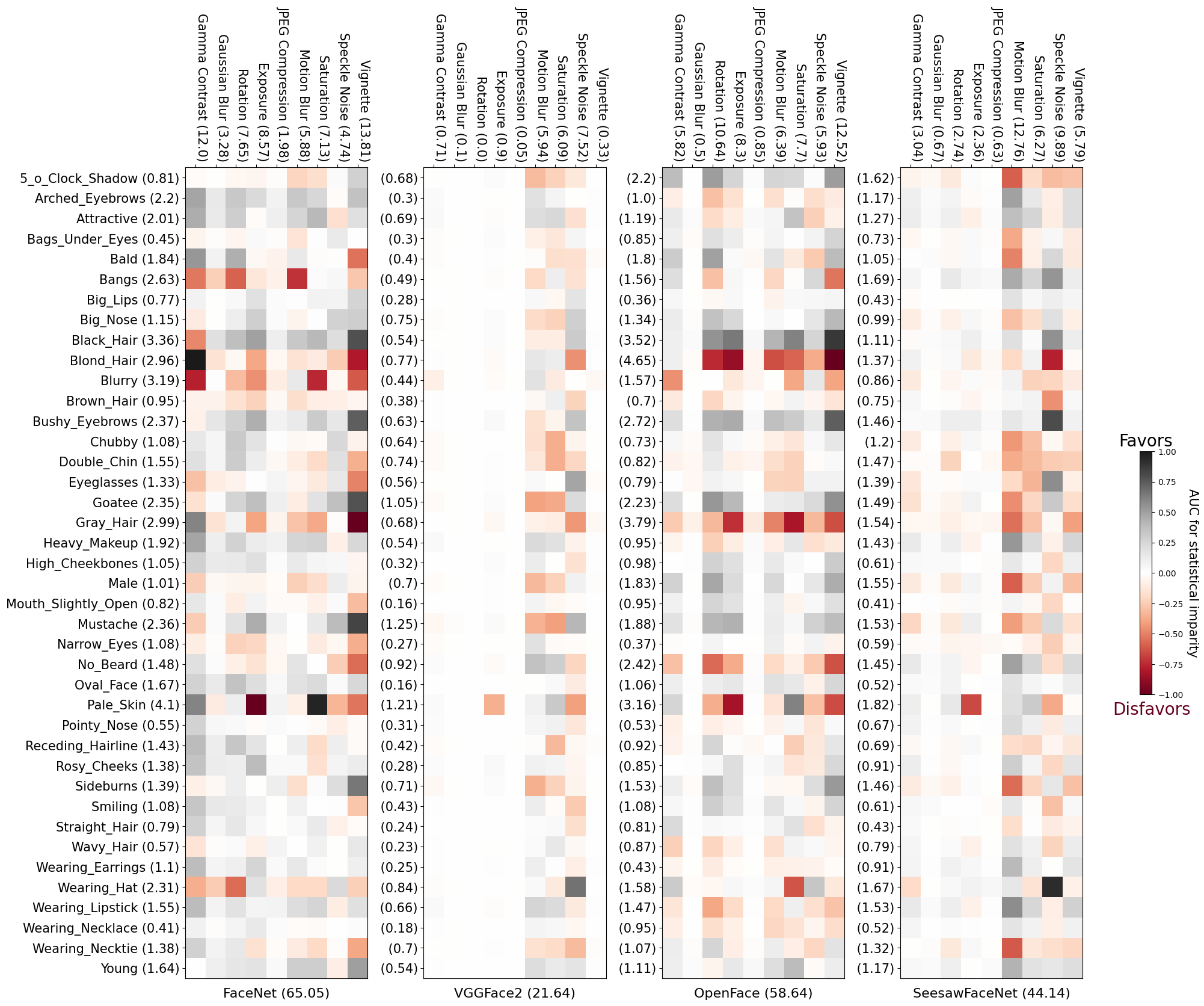}
\caption{Depiction of AUC for statistical imparity curves all attributes without pruning. Note that the numbers in the labels denote the L1 norms.}%
\label{fig:auc_matrix_wopru_fsa}
\centering
\end{figure}

\begin{figure}[htbp]
\centering
\includegraphics[width=\textwidth, height=4in]{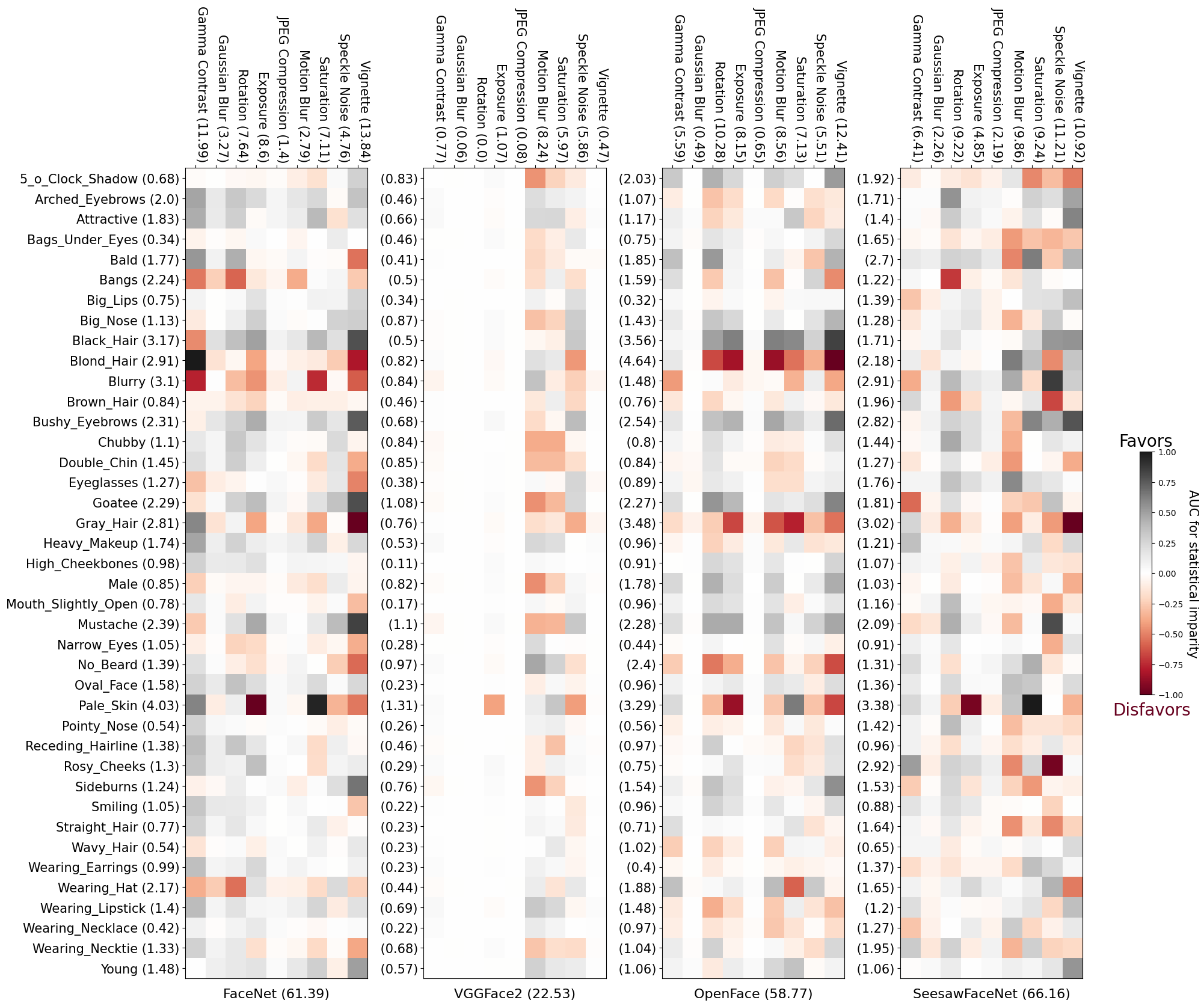}
\caption{Depiction of AUC for statistical imparity curves for the CelebA dataset with pruning. Note that the numbers in the labels denote the L1 norms. Computing matrix L1 norms for the matrices indicates that, on average, VGGFace2 is the most fair (with the lowest L1 norm), and SeesawFaceNet the least fair amongst the models compared.}%
\label{fig:auc_matrix_wpru_fsa}
\centering
\end{figure}

In this section, additional results for self-matching are presented including the complete set of results on all perturbations and attributes for Fair SA, along with the results for experiments using VPSA's pruning step. Fig. \ref{fig:vanilla} shows the AUC matrix for IRCs obtained by VPSA without and with pruning, and we see that the trends obtained with pruning (Fig. \ref{fig:vanilla_b}) are largely similar to those without pruning (Fig. \ref{fig:vanilla_a}). Fig. \ref{fig:compare} compares the plots obtained with and without the pruning step for a sample perturbation and protected subgroup. Figures \ref{fig:auc_matrix_wopru_fsa} and  Fig. \ref{fig:auc_matrix_wpru_fsa} depict the AUC matrices for Fair SA for all attributes without and with pruning respectively.

\begin{figure}[htbp]
\centering
\includegraphics[width=\textwidth, height=2.0in]{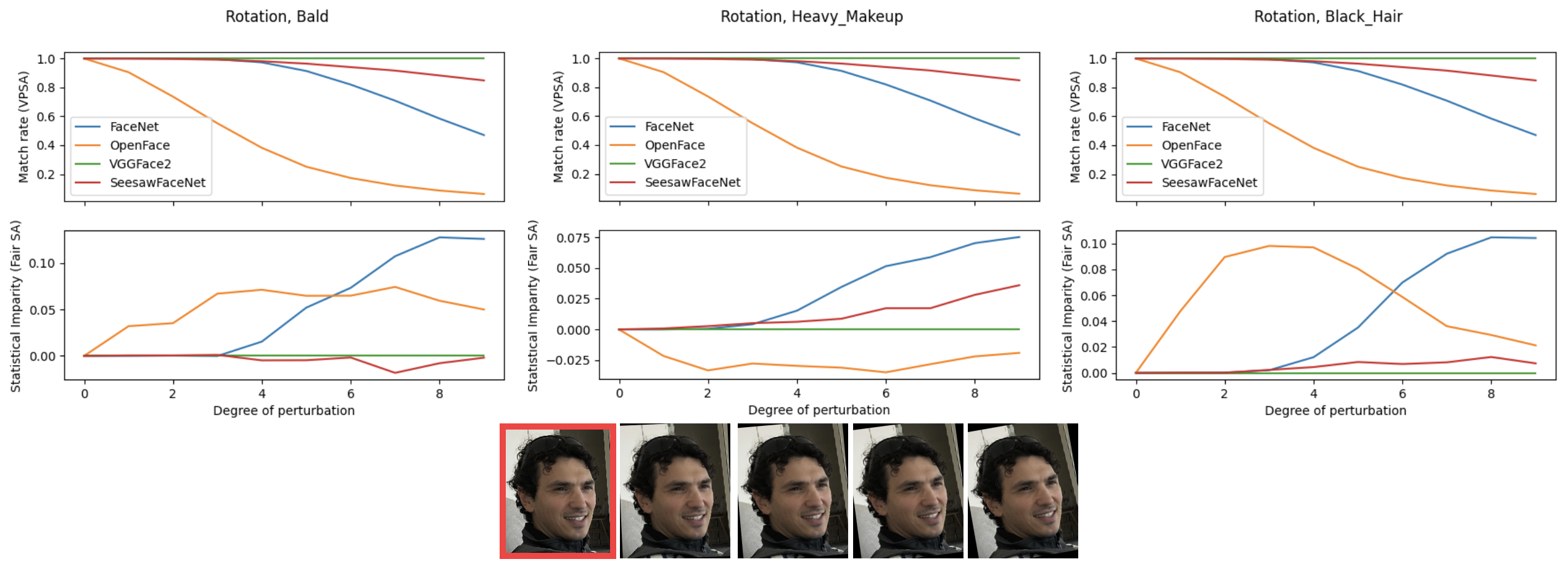}
\caption{Depiction of VPSA and Fair SA curves for the rotation perturbation across sample protected subgroups. We can see that VGGFace2 is absolutely robust and fair since the AUC for the VPSA and Fair SA curves is 1 and 0 respectively.}%
\label{fig:rotation}
\centering
\end{figure}

\end{document}